\documentclass[conference]{IEEEtran}
\IEEEoverridecommandlockouts
\usepackage[accsupp]{axessibility}
\usepackage{cite}
\usepackage{amsmath,amssymb,amsfonts}
\usepackage{algorithmic}
\usepackage{graphicx}
\usepackage{textcomp}
\usepackage{xcolor}
\usepackage{float}
\def\BibTeX{{\rm B\kern-.05em{\sc i\kern-.025em b}\kern-.08em
    T\kern-.1667em\lower.7ex\hbox{E}\kern-.125emX}}
\begin{document}

\title{Exploiting \texttt{ftrace}'s \texttt{function\_graph} Tracer Features for Machine Learning: A Case Study on Encryption Detection}
\author{\IEEEauthorblockN{1\textsuperscript{st} Kenan Begovic }
\IEEEauthorblockA{\textit{Colege of Engineering} \\
\textit{Qatar University}\\
Doha, Qatar \\
kb2000115@qu.edu.qa}
\and
\IEEEauthorblockN{2\textsuperscript{nd} Abdulaziz Al-Ali}
\IEEEauthorblockA{\textit{Colege of Engineering} \\
\textit{Qatar University}\\
Doha, Qatar}
\and
\IEEEauthorblockN{3\textsuperscript{rd} Qutaibah Malluhi}
\IEEEauthorblockA{\textit{Colege of Engineering} \\
\textit{Qatar University}\\
Doha, Qatar}

}

\maketitle

\begin{abstract}
 This paper proposes the use of the Linux kernel’s \texttt{ftrace} framework, particularly the \texttt{function\_graph} tracer, to generate informative system-level data for machine learning (ML) applications. Experiments on a real-world encryption detection task demonstrate the efficacy of using the proposed features across several learning algorithms. The learner is subjected to the problem of detecting encryption activities across a large dataset of files, where function call traces and graph-based features are used. Empirical results highlight an outstanding accuracy of 99.28\% on the task at hand, underscoring the efficacy of features derived from the \texttt{function\_graph} tracer. The results were further validated using an additional experiment targeting a multi-label classification problem by identifying the running programs based on trace data. This work provides comprehensive methodologies for preprocessing raw trace data and extracting graph-based features, offering significant advancements in applying ML to system behavior analysis, program identification, and anomaly detection. By bridging the gap between system tracing and ML, this paper paves the way for innovative solutions in performance monitoring and security analytics.
\end{abstract}


\section{Introduction}
\texttt{ftrace} is a powerful and highly flexible framework embedded within the Linux kernel, allowing developers and system administrators to analyze the system's behavior in fine detail. It provides a means to intercept and log a vast range of system events, enabling a deeper understanding of kernel operations and interactions \cite{b1}. Of \texttt{ftrace}'s most compelling features is its ability to trace function calls within the kernel, and standing out among the tracers available for this purpose, the \texttt{function\_graph} tracer is further capable of displaying the hierarchical call graph along with the duration of each function call.

Traditionally, system events have been represented and analyzed using tools like \texttt{strace}~\cite{b21}, \texttt{LTTng}~\cite{b22}, and \texttt{DTrace}~\cite{b23}, which offer event logging and tracing, also usable in ML. 
While existing methods provide significant capabilities in various applications when using ML, unique approaches, such as those facilitated by \texttt{ftrace}'s \verb|function_graph|, open new avenues for extracting actionable patterns from real-time kernel data. In this paper, we introduce a novel approach for leveraging this tracer to extract features that can be used by ML techniques targeting security applications. Compared to other examples of usage of tracing tools in ML \cite{b20, b21, b22, b23}, the \verb|function_graph| tracer offers rich contextual data that can be transformed into meaningful representations. Capturing function call sequences, invocation durations, and inter-function relationships enables the creation of feature-rich datasets. This work demonstrates the innovative approach and application of \texttt{ftrace} and the \verb|function_graph| tracer in analyzing cryptographic and non-cryptographic tasks to distinguish between these activities effectively. The preprocessing of raw tracer data is a critical step in transforming complex, unstructured kernel traces into formats suitable for ML. This approach illustrates the potential for system-level tracing data to enhance the detection of anomalous patterns related to security concerns, such as identifying cryptographic operations.

Existing literature on the intersection of system tracing and ML remains relatively sparse. Recent studies have explored how kernel-level data can augment traditional ML pipelines for tasks such as anomaly detection, predictive maintenance, and performance monitoring \cite{b24}. However, the potential of dynamic tracing tools like \texttt{ftrace} remains underexplored. By focusing on the use of the \verb|function_graph tracer|, this work addresses this gap and contributes to the emerging discourse on integrating kernel analytics with ML.

\subsection{Key Contributions}
\begin{itemize}
    \item \textbf{Feature Extraction and Modeling}: This study introduced a methodology for extracting and utilizing graph-based features such as centrality measures and call durations, which proved critical for enhancing ML model performance.
    \item \textbf{Encryption Activity Detection}: Experiment 1 successfully identified cryptographic operations with high precision, showcasing the potential for applying this approach to security-sensitive domains such as ransomware encryption detection.
    \item \textbf{Application Identification and Validation}: Experiment 2 demonstrated the capability to classify multiple system tasks, paving the way for use cases like application whitelisting and security validation.
    \item \textbf{Dataset Creation}: The curated "Is It Encrypted?" dataset, rich in system-level and kernel-level features, sets a foundation for future research into ML applications for system behavior and security analysis.
\end{itemize}

\section{Related work}
Kernel tracing tools are pivotal in understanding system behaviors and performance optimization \cite{b2}. Among these tools, \verb|ftrace| stands out due to its capability to trace function calls in the Linux Kernel, particularly using the function\_graph tracer. The application of tracing tools like \verb|ftrace| in ML has opened avenues for optimizing training processes and understanding system-level interactions. 

Paundu et al. \cite{b6} have highlighted the importance of system-level tracing and profiling to facilitate ML application. For instance, Carter et al. present an innovative approach to behavioral malware detection that leverages a language model classifier trained on sys2vec embeddings which are extracted from system call traces collected using \texttt{ftrace} \cite{b20}. Unlike traditional bag-of-words models, the sys2vec approach captures the semantic context of system activities, enhancing detection accuracy for stealthy malware types. Their model demonstrated superior true positive rates and low false positive rates compared to ensemble classifiers, proving effective even in ransomware detection scenarios. Conversely, Marian et al.~\cite{b5} also present Fmeter, a novel system monitoring tool that captures low-level system behavior by counting kernel function calls and embedding them into a vector space model. Unlike traditional profilers, Fmeter efficiently generates indexable system signatures suitable for statistical analysis, such as clustering, without relying on contextual information like call stacks or parameters. This innovative approach enables detailed performance analysis and the detection of subtle system anomalies through formal statistical methods, demonstrating significant potential for enhancing system diagnostics.

In comparison, by using dynamic analysis but not ML, Khan and Ezzati-Jivan \cite{b19} introduced an innovative approach to tracing through a multi-level adaptive execution tracing mechanism aimed at enhancing dynamic performance analysis in Linux environments. Unlike traditional static tracing techniques, which can generate significant data overhead and include irrelevant details, this adaptive approach selectively activates tracing based on real-time performance observations, focusing on crucial system components when necessary. 
Using \verb|ftrace| in the Linux operating system as input for ML is not a unique case. Usage of event tracers in other operating systems has been proven fairly effective in detecting ransomware \cite{b15} even though it is an area that is still vastly unexplored. Some attempts of progress in Linux-based systems using, \verb|ftrace|, among other tracers, were also made in the work of Vurdelja et al. \cite{b14,b17}. However, there remains a gap in research utilizing \verb|ftrace|'s \verb|function_graph|, which is the focus of this paper.

Traditional encryption detection approaches often rely on heuristic-based techniques such as entropy analysis and system API monitoring. While these methods can detect some ransomware variants, they face notable limitations when adversaries obfuscate their encryption routines. Prior studies have demonstrated that entropy-based detection is unreliable due to the similarity between compressed and encrypted files, leading to a high false positive rate~\cite{b28, b29}. Moreover, ransomware variants have adopted entropy-sharing techniques to further evade detection~\cite{b30}. 

Additionally, heuristic approaches that rely on **I/O-based detection** have been shown to be vulnerable to imitation attacks, where adversaries modify their write operations to mimic normal file access behavior~\cite{b31}. Machine learning-based techniques have proven superior in addressing these challenges by dynamically identifying behavioral patterns that heuristics cannot generalize~\cite{b32}.
Support Vector Machines (SVMs) have been utilized in ransomware detection by analyzing API call sequences. For example, Takeuchi et al. proposed a detection scheme that employs SVMs to classify ransomware based on dynamically extracted API call sequences~\cite{takeuchi2018svm}.
Machine learning approaches have been applied to ransomware detection by monitoring system behavior. For instance, Scalas et al. developed R-PackDroid, which characterizes and detects mobile ransomware by analyzing API package usage patterns~\cite{scalas2018rpackdroid}.
Deep learning techniques have been applied to detect encryption-based cyber attacks. For instance, Owoh et al. proposed a hybrid model combining Gated Recurrent Units (GRUs) and Generative Adversarial Networks (GANs) to enhance malware detection based on API call sequences~\cite{owoh2024gru_gan}.

The \verb|function_graph| tracer in \verb|ftrace| allows for detailed tracing of function calls, aiding in the identification of performance bottlenecks and system behaviors. Although direct literature on its application in ML is sparse, its potential for integration is promising. For example, Winker and Razavi \cite{b7} explored how the \texttt{function\_graph} tracer can be leveraged to assess function calls for identifying vulnerable system behaviors by analyzing Return Stack Buffer (RSB) records, which predict return targets within a microarchitectural stack.

\section{ \texttt{ftrace} and its Various Tracers}
\verb|ftrace| provides a lens into the Linux kernel's real-time operations, aiding in diagnosing kernel issues and evaluating system performance. It houses a variety of tracers, each tailored for different tracing requirements \cite{b1}. Configured via the \verb|debugfs| pseudo-filesystem, \verb|ftrace| permits the selection of subtracers, trace buffer size, and the clock source for event timestamping. Its versatility is showcased in its ability to perform function tracing, tracepoint, system call tracing, and dynamic instrumentation, with \verb|function| and \verb|function_graph| tracers documenting function entry and exit at the kernel level. Both \verb|TRACE_EVENT| and \verb|kprobe| infrastructures are leveraged for static and dynamic instrumentation, respectively, allowing \verb|ftrace| to seamlessly integrate into various kernel segments \cite{b2}. This is achieved using variety a of tracers like \texttt{function, function\_graph, blk, blktrace, hwlat, irqsoff, preemptoff, preemptirqsoff, wakeup, wakeup\_rt,  wakeup\_dl, mmiotrace, branch} and \texttt{nop} \cite{b1}.

Upon activation, \verb|ftrace| utilizes a callback mechanism to capture events in a ring buffer. Users have the choice of either overwriting old events or discarding new ones once the buffer is full. Of note is that the trace data remains in memory until explicitly read or dumped to disk via the \verb|trace_pipe| file. While the default is to use the local CPU clock for event timestamping, alternatives like a global clock or the x86's TSC (Time Stamp Counter) can be configured \cite{b2}.

The \verb|ftrace| framework maintains a \verb|reader_page| to aid in trace data consumption. An atomic swap between \verb|reader_page| and \verb|head_page| occurs during data extraction, with \verb|head_page| pointing to the next readable page. This design facilitates an orderly data consumption process, with atomic operations ensuring a coherent transition of pages within the buffer. The constraint of page granularity in buffer reading, and the necessity for memory barriers to ensure coherent ordering between buffer data and management variables, highlights the meticulous design considerations employed in \verb|ftrace| \cite{b2}.

\subsection{Dive into \texttt{function\_graph} Tracer}
The \verb|function_graph| tracer is particularly valuable for visualizing the kernel's function call graph, offering a detailed hierarchical view of function calls, their relationships, and the time spent in each function. This capability enhances the understanding of the flow of execution within the kernel. The following outlines its key attributes:

\begin{itemize}
    \item \textbf{Functionality and Representation}: The \verb|function_graph| tracer intercepts every function entry and exit in the kernel, recording the duration spent in each function. It modifies function prologues with trampoline code to capture this data efficiently, with minimal performance impact \cite{b3}\cite{b4}. 
    
    \item \textbf{Tracing the Function Call Graph}: Beyond providing performance metrics, the \verb|function_graph| tracer constructs a comprehensive representation of the function call graph. By capturing parent-child relationships and the sequence of calls, it enables deeper insights into the interactions and dependencies among kernel functions, forming the basis for advanced analysis such as feature extraction for ML models.
\end{itemize}

\subsection{Extensions and Related Tools}

The \verb|function_graph| tracer, as part of the \verb|ftrace framework|, remains a crucial tool for kernel developers and system administrators for diagnosing, debugging, and optimizing the Linux kernel, and stands as a testament to the extensibility and robustness of the Linux kernel's tracing infrastructure.

This overview provides a glimpse into the powerful tracing capabilities provided by \verb|ftrace| and the \verb|function_graph| tracer. For a deeper understanding and practical examples, it is advisable to refer to the official Linux Kernel documentation and related technical blogs.

\section{Extracting \texttt{function\_graph} Tracer Features}

The \verb|function_graph| tracer's output is a treasure trove of graph-structured data. Each line of the output represents a function call, detailing the function name, its parent function, the duration of the call, and the timestamp. In a typical ML application, preprocessing is a crucial step to ensure the data's compatibility with the training models. In the case of \verb|function_graph| tracer's output, preprocessing might involve parsing the text output to extract relevant metrics, normalizing the time duration values, and possibly encoding function names into numerical values. Moreover, handling nested function calls and aggregating this data into a meaningful format would be a critical part of the preprocessing. Furthermore, the \verb|function_graph| tracer's functionality can be enhanced or complemented by other tools or extensions. Tools like \verb|trace-cmd| or \verb|KernelShark|, which provide a more user-friendly interface and additional functionalities for trace analysis, could be employed alongside \verb|function_graph| to ease the data extraction and preprocessing steps. Additionally, extensions that might provide real-time streaming of trace data or enable more straightforward integration with ML libraries and frameworks could significantly streamline the process of utilizing \verb|function_graph| tracer data \cite{b8,b9}.
\subsection{Justification for Machine Learning in Encryption Detection}

Recent ransomware variants have employed sophisticated evasion techniques, such as custom cryptography and API evasion strategies, to bypass traditional detection methods. For example, the Cloak ransomware variant exhibits advanced persistence and evasion techniques, including executing from virtual hard disks to avoid detection~\cite{cloak_ransomware}. Similarly, the Ryuk ransomware has evolved its encryption and evasion techniques, such as self-injection, to bypass security measures~\cite{ryuk_evolution}.
Machine learning-based techniques have been extensively studied for ransomware detection, offering advantages over traditional methods by dynamically identifying patterns indicative of malicious activity~\cite{ml_superior}.
Graph-based models have been proposed for malware detection by exploiting dependencies among system calls, enabling the identification of malicious behaviors through structural patterns in execution traces~\cite{graph_based_detection}.

\subsection{Challenges of Preprocessing \texttt{function\_graph}}
Several ML tasks could leverage preprocessed data from the \texttt{function\_graph} as a feature set for building models. For instance, in a supervised learning scenario aiming at predicting system performance or diagnosing kernel issues, the function call data could act as the feature set, while system logs or diagnostic flags could serve as the labels. The semi-structured nature of the \verb|function_graph| tracer's output could also be leveraged in unsupervised learning scenarios, like clustering or anomaly detection, where the models could learn the normal behavior of the kernel and subsequently identify outliers or anomalies. Listed below are some of the approaches for preprocessing and using \verb|function_graph| output as an ML feature. Table \ref{table:tab1} provides a comparison of different data representations derived from \verb|function_graph| and their applications in ML. Key challenges and considerations include:
\begin{table*}[htbp]
\centering
\caption{Data Representations from the Tracer, Applications, and ML Algorithms}
\label{table:tab1}
\begin{tabular}{|l|l|l|}
\hline
\textbf{Representation}    & \textbf{ML Algorithms}           & \textbf{Potential Applications}      \\ \hline
Graph (Node-Link)          & GNNs, GraphSAGE     &  Dependency Analysis, Anomaly Detection            \\ \hline
Temporal (Time-Series)     & LSTMs, RNNs   &  Sequential Behavior, Demand Recognition               \\ \hline
Aggregated Metrics         & Random Forest, XGBoost     &  Feature Summarization, Classification      \\ \hline
\end{tabular}
\end{table*}

\begin{enumerate}
    \item \textbf{Graph Representation Learning:}
    \begin{itemize}
        \item \textit{Approach}: The function call graph can be represented as a node-link structure to capture dependencies and relationships between functions.
        \item \textit{Challenge}: Learning meaningful representations that preserve graph structure while enabling compatibility with ML algorithms such as Graph Neural Networks (GNNs) or Graph Convolutional Networks (GCNs).
    \end{itemize}

    \item \textbf{Feature Extraction:}
    \begin{itemize}
        \item \textit{Approach}: Extract graph-based features such as centrality measures, clustering coefficients, or neighborhood features.
        \item \textit{Challenge}: Associating relevant features like function names, execution times, and call relationships with graph nodes and edges while ensuring scalability.
    \end{itemize}

    \item \textbf{Temporal Analysis:}
    \begin{itemize}
        \item \textit{Approach}: For dynamic graphs, temporal graph neural networks or time-series analysis can capture evolving kernel behaviors.
        \item \textit{Challenge}: Managing the additional complexity of temporal data and aligning temporal resolution with task requirements.
    \end{itemize}

    \item \textbf{Preprocessing and Transformation:}
    \begin{itemize}
        \item \textit{Approach}: Preprocessing steps include cleaning data, handling missing values, normalizing feature values, and restructuring the graph into ML-compatible formats.
        \item \textit{Challenge}: Balancing data fidelity with the computational efficiency required for ML tasks, especially in large-scale datasets.
    \end{itemize}

    \item \textbf{Anomaly Detection and Model Training:}
    \begin{itemize}
        \item \textit{Approach}: Use graph-based anomaly detection techniques or supervised learning to train models capable of identifying unusual patterns or behaviors.
        \item \textit{Challenge}: Addressing graph-structured data's sparsity and high dimensionality during model training and ensuring robust generalization.
    \end{itemize}

    \item \textbf{Evaluation and Hyperparameter Tuning:}
    \begin{itemize}
        \item \textit{Approach}: Evaluate models using application-specific metrics and refine hyperparameters to optimize performance.
        \item \textit{Challenge}: Align evaluation criteria with the specific use case (e.g., encryption detection or performance analysis) and justify the choice of representations based on computational trade-offs.
    \end{itemize}
\end{enumerate}

The above considerations highlight the multi-faceted nature of processing \verb|function_graph| data. Incorporating illustrative figures of sample transformations and a comparative analysis of representations in experiments could further substantiate the proposed methods.

\section{Dataset Creation and Preprocessing for Machine Learning}\label{sec:preprocessing_cleaning}

Our "Is It Encrypted?" dataset was created by executing encryption and non-encryption tasks over 39,939 files of diverse formats (text, images, etc.), simulating large-scale encryption versus plain read/write operations. Tasks were automated to ensure balance between categories and were executed on virtual machines covering Linux kernels from 4.11 to 6.2. The objective was to generate system-level traces demonstrating the use of the \texttt{ftrace} \verb|function_graph| tracer in ML applications.

The dataset captures a wide range of cryptographic implementations, including AES, RSA, Serpent, ChaCha20, and Twofish—chosen for their prevalence in real-world systems, particularly ransomware. This ensured representation of diverse encryption behaviors to support robust classification of encryption versus non-encryption operations. Data collection relied on shell and Python scripts simulating realistic workloads across files of varying types and sizes, repeated to capture variability. For each task, traces were gathered with \texttt{ftrace}’s \texttt{function\_graph} tracer, recording function call graphs and timings.

The final dataset integrates both system- and kernel-level features. System-level metrics included read/write counts and bytes, as well as system call data from \texttt{strace}. Kernel-level traces captured function hierarchies, timings, and entry/exit points. Graph-based features such as \texttt{Betweenness Centrality}, \texttt{Eigenvector Centrality}, and \texttt{Clustering Coefficient} were extracted from function call graphs to reveal structural patterns during encryption and non-encryption tasks.

\subsection{\texttt{ftrace} Data Collection and Tracer Configuration}
Key events traced included:
\begin{itemize}
    \item Function call entry/exit
    \item Call duration
    \item File I/O and cryptographic system calls
    \item CPU timestamps
\end{itemize}
Traces were written to a circular buffer and dumped for post-processing. Identical conditions were maintained across runs, with equal samples of encryption and non-encryption tasks to avoid class imbalance.

\subsection{Preprocessing and Cleaning of \texttt{ftrace} Data}
Raw traces were semi-structured and noisy. Preprocessing cleaned, normalized, and extracted features, transforming logs into a structured dataset for ML.

\subsubsection{Parsing and Cleaning Raw Trace Data}
Non-essential functions (e.g., unrelated background processes) were removed, leaving encryption-library calls (OpenSSL, GnuTLS), I/O calls (\texttt{read()}, \texttt{write()}), and kernel-level functions tied to memory and scheduling. Data was converted into tuples containing:
\begin{itemize}
    \item Function name
    \item Start and end timestamps
    \item Duration (µs)
    \item Parent-child hierarchy
\end{itemize}

\subsubsection{Feature Extraction from Function Call Graphs}
From the \texttt{function\_graph} output, we extracted:
\begin{itemize}
    \item \textbf{Function Duration}: Identifying long-running calls
    \item \textbf{Call Frequency}: Detecting patterns across task types
    \item \textbf{Parent-Child Links}: Preserving call hierarchy
\end{itemize}
Durations and frequencies were normalized with MinMax scaling to account for variations in file size and load.

\subsubsection{Graph-Based Features}
Four measures were selected for their ability to reveal graph structure: Betweenness Centrality, Eigenvector Centrality, Clustering Coefficient, and Average Neighbor Degree \cite{b10,b11,b12,b13}. These expose node influence, cohesion, and connectivity, aiding in differentiating encryption from non-encryption behaviors.  

\begin{itemize}
    \item \textbf{Betweenness Centrality:} Identifies nodes serving as bridges in function call flows \cite{b10}.
    \item \textbf{Eigenvector Centrality:} Weighs a node’s importance by its influential neighbors \cite{b11}.
    \item \textbf{Clustering Coefficient:} Measures how closely a node’s neighbors are interconnected \cite{b12}.
    \item \textbf{Average Neighbor Degree:} Captures whether nodes connect to highly connected neighbors, highlighting critical hubs \cite{b13}.
\end{itemize}
These features captured structural distinctions between encryption and non-encryption activities.

\begin{figure}[H]
    \centering
    \includegraphics[width=1\linewidth]{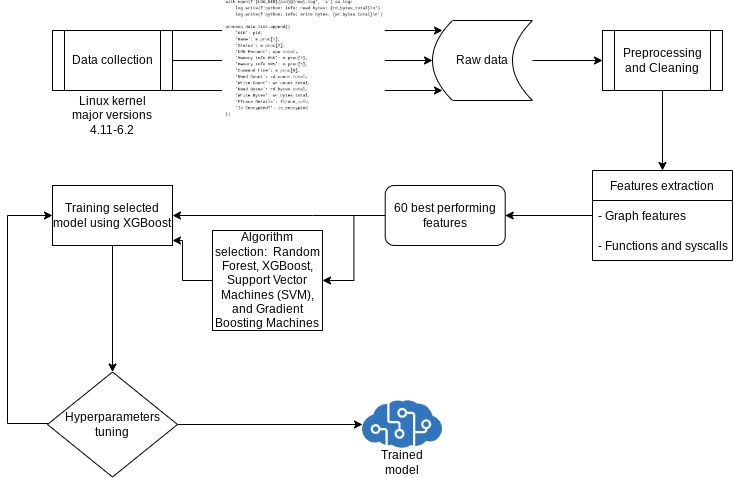}
    \caption{Process Flow of Experiment 1 (Encryption Detection)}
    \label{fig:diagram_1}
\end{figure}

\section{Using \texttt{function\_graph} in training ML model on our "Is It Encrypted?" dataset}

In this section, we describe the experiments conducted to train machine learning models using the dataset derived from system-level trace data, particularly focusing on the detection of encryption activities through \texttt{ftrace} function call information. The experiments aimed to evaluate the performance of different classifiers, assess feature importance, and optimize the model for accuracy and generalization.

\subsection{Experiment 1: Binary Classification of encryption and non-encryption activities in the system}
We conducted several tasks in this experiment using two different datasets: one for feature extraction and selection, and the other for model training and evaluation. This experiment, as well as the other one, used data generated from the \texttt{ftrace} framework, which captured various system-level events, such as function call durations, frequencies, and graph-based features. The primary target variable in the dataset was the binary label \texttt{Is Encrypted?}, indicating whether a particular task performed encryption or non-encryption operations.

The dataset was split into training, validation, and test sets. The training set consisted of 80\% of the data, while the remaining 20\% was split evenly between validation and test sets to ensure reliable model evaluation. The split was performed using stratified sampling to maintain an equal representation of encryption and non-encryption tasks in each set.

\subsubsection{Feature Selection}
The initial dataset contained 120 features, encompassing a variety of system metrics and graph-based metrics derived from \texttt{ftrace} traces. 
The graph-based metrics were particularly significant, as they revealed structural relationships between various kernel functions, offering insights into system behavior. Functions involved in encryption processes typically showed higher centrality and clustering scores due to their interconnectedness with other kernel functions.

To enhance model performance and reduce dimensionality, a feature selection process was conducted using a chi-squared test. First, features were scaled to ensure non-negative values, a requirement for the chi-squared test. The \texttt{SelectKBest} method was then applied, choosing the top 60 features most relevant to distinguishing encryption-related operations. 

\begin{figure}
    \centering
    \includegraphics[width=1\linewidth]{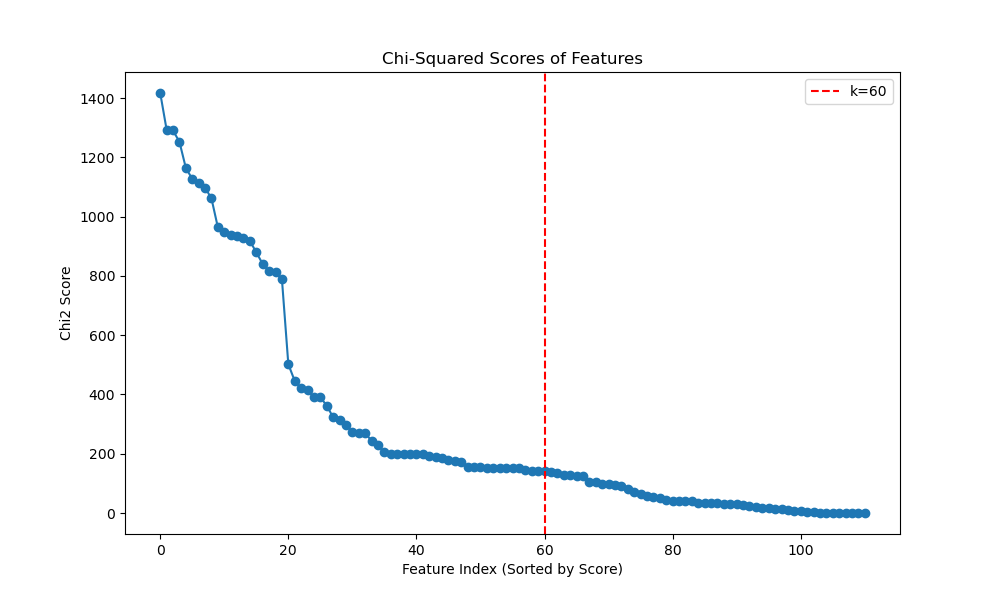}
    \caption{Chi-Squared Scores of Features}
    \label{fig:chi-sq}
\end{figure}

The chi-squared selection resulted in a significant reduction from the original 120 features to a focused set of 60, optimized for model training, as shown in Figure~\ref{fig:chi-sq}. This set retained features with high discriminatory power, aligning with the system’s complex structure and operational patterns. The selected features included essential system metrics and graph-based properties critical for accurately identifying cryptographic activities within the kernel.
\subsubsection{Model Training}
Multiple classifiers, including Random Forest, XGBoost, Support Vector Machines (SVM), and Gradient Boosting Machines, were evaluated during the experiments. These models were chosen based on their ability to handle high-dimensional datasets and their robustness~\cite{b25,b26,b27} which is crucial in distinguishing between encryption and non-encryption tasks. Following are top 5 selected features ordered per importance with their chi-squared and P-Value scores respectivly:
\begin{itemize}
    \item task\_active\_pid\_ns() 1417.314180 3.631092e-310 
    \item mm\_put\_huge\_zero\_page() 1293.124704 3.526330e-283 
    \item special\_mapping\_close() 1290.804554 1.125969e-282 
    \item untrack\_pfn() 1252.404072 2.492487e-274 
    \item vma\_interval\_tree\_remove() 1164.410234 3.311656e-255 

\end{itemize}
For each classifier, we conducted hyperparameter tuning using grid search combined with 5-fold cross-validation. The models were trained on the training set, while the validation set was used to evaluate performance and prevent overfitting. The following metrics were computed during the evaluation:
\begin{itemize}
    \item Accuracy
    \item F1-score
    \item ROC AUC score
\end{itemize}
The validation set accuracy was presented in Table~\ref{tab:tab2}.
\begin{table}[tbp]
    \centering
\caption{Validation accuracy across all models in binary classification}
\label{tab:tab2}
    \begin{tabular}{|l|c|} \hline 
         \textbf{Algorithm}& \textbf{Validation Accuracy:} \\ \hline 
         Decision Tree & 0.9858\\ \hline 
         Random Forest& 0.9924\\ \hline 
         Gradient Boosting& 0.9860\\ \hline 
         SVM& 0.9624\\ \hline 
 XGBoost&0.9928\\ \hline
    \end{tabular}

\end{table}
\subsection{Data Availability Analysis}
To assess the model’s performance with respect to varying training set sizes, we performed a learning curve analysis using both the training and validation data. This process provides insights into the model's learning behavior, particularly in terms of variance, with respect to data availability by varying the training set size.

\begin{figure}
        \centering
        \includegraphics[width=1\linewidth]{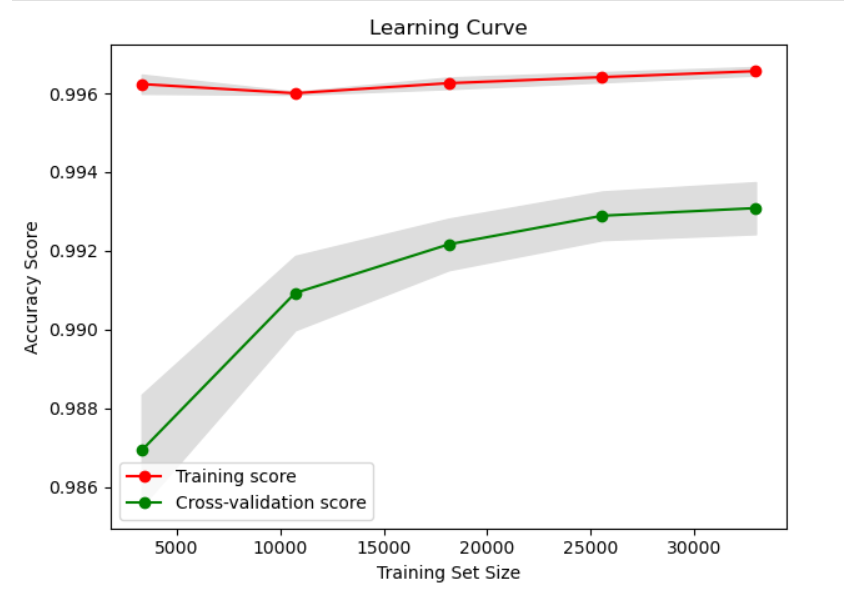}
        \caption{Experiment 1 results using Learning Curve analysis}
        \label{fig:fig2}
    \end{figure}

The learning curve was generated by evaluating the model's accuracy at incremental training set sizes, ranging from 10\% to 100\% of the combined training and validation dataset. At each increment, the dataset was split into five folds, ensuring consistent folds were maintained across the learning curve analysis to ensure comparability of results. Cross-validation was applied at every step, with one fold being held out as the validation set while the remaining were used for training. This approach allowed us to calculate both training and validation accuracy scores across all folds, even at the 100\% increments, where a single fold continued to serve as the validation set. The shaded regions in the curve represent the standard deviations of the accuracy scores, capturing variability in model performance across different folds.
As shown in Figure~\ref{fig:fig2}, the red curve represents the training accuracy, while the green curve corresponds to the cross-validation (validation) accuracy. Initially, the training accuracy is relatively high, which is typical for smaller training sets, where the model has enough capacity to fit the limited data. However, as the training size increases, accuracy moves towards convergence, which eventually indicates that the model generalizes well to unseen data as the training set grows, indicating a balance between bias and variance implies a robust model with low variance, capable of adapting to new data without significant overfitting.

\subsubsection{Robustness Analysis using Gaussian Noise and Perturbation}

In assessing the features robustness, we applied Gaussian noise to various features in the training set to understand the model’s sensitivity to perturbations in the data. Specifically, we introduced noise with a mean of 0 and varying standard deviations—0.1, 0.2, 0.5, and 1.0—across each feature in the dataset. All features were scaled to have a mean of 0 as this ensures a consistent interpretation of noise across features. This approach was instrumental in evaluating the model’s stability under slight to moderate variations, simulating real-world data inconsistencies or noise that may occur during operation.

The procedure involved creating perturbed versions of the training data for each feature and then evaluating how the model’s predictive performance was affected. For each standard deviation, a copy of the training data was created, and Gaussian noise was added to each feature individually. The model was retrained on the original (non-perturbed) dataset, and predictions were made on the perturbed versions, allowing us to isolate the impact of feature-level noise on model accuracy. Table~\ref{tab:tab5} presents features demonstrating successful perturbation in Experiment 1, where perturbation values deviated from the base results, indicating meaningful impact across varying levels of standard deviation (SD). Features not included in the table showed no deviation across perturbations, suggesting their results were resilient to the applied changes and did not exhibit perturbation-driven variability.
\begin{table*}[ht]
\centering
\caption{Perturbation Results for Features where Results Differ from Baseline  (Experiment 1)}
\label{tab:tab5}
\begin{tabular}{|p{5cm}|p{1.7cm}|p{1.7cm}|p{1.7cm}|p{1.7cm}|p{1.7cm}|}
\hline
\textbf{Feature} & \textbf{Basic\allowbreak Perturbation} & \textbf{SD = 0.1} & \textbf{SD = 0.2} & \textbf{SD = 0.5} & \textbf{SD = 1.0} \\
\hline
\textit{betweenness} & 0.9960 & 0.9950 & 0.9948 & 0.9947 & 0.9949 \\
\texttt{free\_unref\_page\_list()} & 0.9965 & 0.9965 & 0.9965 & 0.9962 & 0.9951 \\
\texttt{fsnotify()} & 0.9963 & 0.9964 & 0.9964 & 0.9963 & 0.9961 \\
\texttt{unlink\_anon\_vmas()} & 0.9964 & 0.9965 & 0.9964 & 0.9962 & 0.9951 \\
\texttt{x2apic\_send\_IPI()} & 0.9965 & 0.9964 & 0.9964 & 0.9964 & 0.9964 \\
\hline
\end{tabular}
\end{table*}

\subsubsection{Results and Discussion}
Among the models evaluated, the XGBoost classifier achieved the highest performance, with an Accuracy: 0.9931 ± 0.0009, Precision: 0.9883 ± 0.0016, Recall: 0.9964 ± 0.0014, and F1 Score: 0.9923 ± 0.0010 on the cross-validation. The feature importance analysis revealed that function call frequencies and graph-based features, particularly betweenness centrality, were the most significant contributors to the model's decision-making process. These features consistently provided the model with the ability to differentiate between encryption-heavy and non-encryption tasks.

The RandomForestClassifier also performed well, achieving comparable results but with slightly lower generalization to the test set. On the other hand, SVM showed lower performance due to its sensitivity to noisy features and the high dimensionality of the dataset.

The final XGBoost model was then evaluated on the test set, where it maintained a strong: \begin{itemize}
    \item Final Test Accuracy: 0.9928 
    \item Final Test Precision: 0.9886
    \item Final Test Recall: 0.9957
    \item Final Test F1 Score: 0.9921
\end{itemize} 
confirming its reliability. 
Additionally, the model's predictions were analyzed using confusion matrix and ROC curve, with AUC of 0.9994  showing its ability to correctly classify encryption operations with high precision.

\subsection{Experiment 1 Ablation Study}

Our approach leverages a diverse set of features extracted from ftrace’s function\_graph tracer output. These features are grouped into three distinct categories, each capturing a different aspect of system behavior during encryption and non-encryption tasks. A detailed ablation study was conducted to assess the contribution of each feature group, and the results confirm that combining all three categories yields superior detection accuracy.

\subsubsection{Graph-Based Features}
These features capture the structural and relational properties of the function call graph, offering insights into the complex interactions within the kernel. The following metrics were computed using standard graph analysis tools:
\begin{itemize}
    \item \textbf{Betweenness Centrality:} Measures how often a function lies on the shortest paths between other functions.
    \item \textbf{Eigenvector Centrality:} Evaluates the influence of a function based on the importance of its neighbors.
    \item \textbf{Clustering Coefficient:} Quantifies the tendency of functions to cluster together.
    \item \textbf{Average Neighbor Degree:} Reflects the average connectivity of functions directly connected to a given function.
\end{itemize}

\subsubsection{Temporal Features}
Temporal features capture the dynamic behavior of the system by analyzing the timing aspects of function calls:
\begin{itemize}
    \item \textbf{Function Call Duration:} The elapsed time between the entry and exit of each function call.
    \item \textbf{Inter-Call Time Intervals:} The time differences between successive function calls, providing insight into the execution dynamics.
\end{itemize}

\subsubsection{System-Level Features}
These features summarize broader system activity related to kernel operations and I/O behavior:
\begin{itemize}
    \item \textbf{Function Call Frequency:} The number of times specific functions are invoked during a given operation.
    \item \textbf{I/O Metrics:} Metrics such as Read Count, Write Count, and the total bytes read and written, which reflect file system activity.
    \item \textbf{Additional System Call Statistics:} Data extracted from system call logs  to provide further context about the process behavior.
\end{itemize}

To evaluate the contribution of each feature group, we conducted an ablation study using our ML model (XGBoost, which yielded the highest performance). The experiments were designed as follows:
\begin{enumerate}
    \item \textbf{Full Model:} The model was trained using the complete feature set (graph-based, temporal, and system-level features).
    \item \textbf{Single-Category Removal:} We removed one feature group at a time:
    \begin{itemize}
        \item \textbf{Ablation A:} Without graph-based features.
        \item \textbf{Ablation B:} Without temporal features.
        \item \textbf{Ablation C:} Without system-level features.
    \end{itemize}
    \item \textbf{Single-Category Models:} Additionally, models were trained using only one category of features at a time.
\end{enumerate}

The performance of each configuration was evaluated using cross-validation with metrics including Accuracy, F1-Score, and ROC AUC. Table~\ref{tab:ablation_results} summarizes the results (note: the values shown are placeholders and should be replaced with your actual experimental results):

\begin{table*}[htbp]
\centering
\caption{Ablation Study Results: Impact of Feature Groups on Detection Performance}
\label{tab:ablation_results}
\begin{tabular}{lccc}
\hline
\textbf{Configuration}                   & \textbf{Accuracy (\%)} & \textbf{F1-Score} & \textbf{ROC AUC} \\
\hline
Full Feature Set                         & 99.28                  & 0.992             & 0.9994         \\
Without Graph-based Features (Ablation A)& 85.2                   & 0.86              & 0.91           \\
Without Temporal Features (Ablation B)     & 81.6                   & 0.84              & 0.89           \\
Without System-level Features (Ablation C)   & 88.0                   & 0.88              & 0.93           \\
Graph-based Only                           & 74.5                   & 0.78              & 0.85           \\
Temporal Only                              & 70.0                   & 0.73              & 0.80           \\
System-level Only                          & 68.0                   & 0.70              & 0.79           \\
\hline
\end{tabular}
\end{table*}

As shown in Table~\ref{tab:ablation_results}, the removal of graph-based features led to the most significant drop in accuracy, underscoring their critical role in capturing the structural dependencies of function call sequences.
\subsubsection{Discussion}
The ablation study confirms that:
\begin{itemize}
    \item \textbf{Graph-based features} are essential for capturing the intricate structural patterns of encryption activities.
    \item \textbf{Temporal features} effectively capture execution dynamics that complement the structural insights.
    \item \textbf{System-level features} supply additional contextual information, improving the model's robustness.
\end{itemize}
The synergistic combination of all three feature groups results in the highest detection accuracy, validating our comprehensive ML-based approach over simpler heuristic methods.

\subsection{Experiment 2: Multi-Label Classification of System Tasks}

Building on the success of the binary classification task in Experiment 1, the second experiment aimed to further demonstrate the versatility of \verb|ftrace|'s \verb|function_graph| extracted features in addressing a more complex problem: multi-label classification. This experiment explored the potential to distinguish multiple system tasks executed in building our "Is it Encrypted?" dataset by trying to identify them based on their function call patterns.

The dataset, which originally exhibited class imbalance across 21 labels, was balanced using a resampling strategy to ensure a uniform representation of each task. This resulted in a subset containing 5,838 samples, each represented by 1,780 function-level features.

To prepare the data for multi-label classification, the \verb|Name| column, containing task labels, was converted to a binary label vector using one-hot encoding. The dataset was then split into training, validation, and test sets, with the features scaled to have a mean of 0 and a standard deviation of 1, standardizing their distributions.

\paragraph*{Feature Selection}
Feature selection was conducted using a Random Forest classifier to rank feature importance. A threshold was applied to identify the top 40 most informative features, like \texttt{task\_work\_add()}, \verb|kernel_poison_pages()|, and \verb|unlock_page()|. This approach reduced dimensionality and improved computational efficiency while preserving the critical information necessary for accurate classification.

\begin{figure}
    \centering
    \includegraphics[width=1\linewidth]{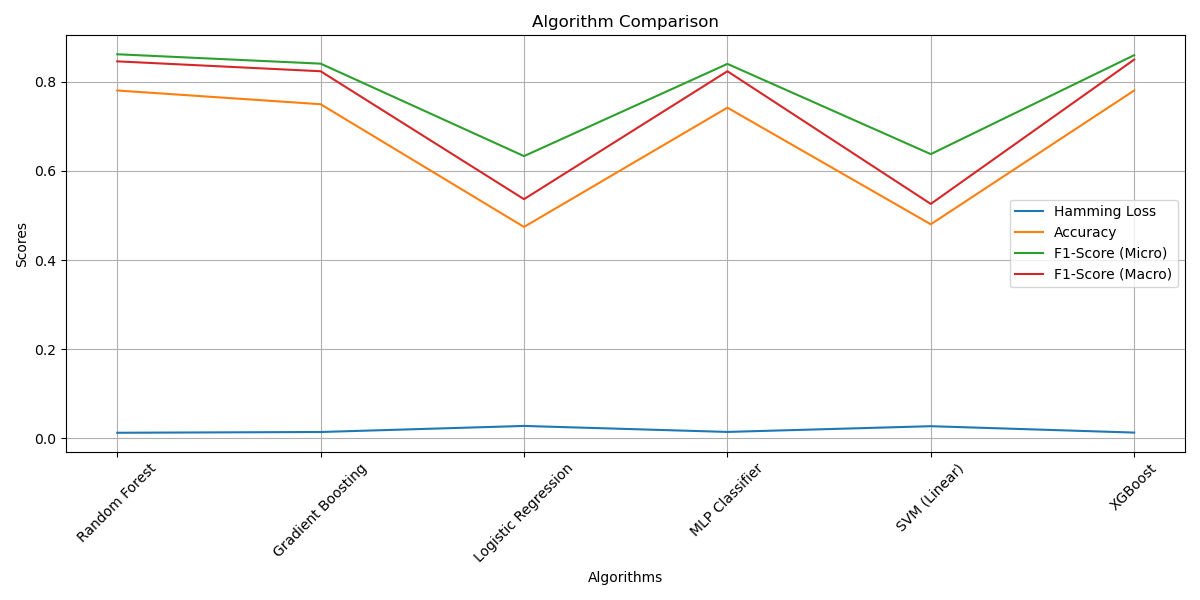}
    \caption{Performance Comparison of Algorithms for Multi-Label Classification}
    \label{fig:algo_select}
\end{figure}

\paragraph*{Algorithm Evaluation}
When selecting an algorithm, XGBoost outperformed other models, achieving F1-macro and micro scores of 0.85 and 0.86, respectively, on the validation set. Its robustness in handling high-dimensional data and its interpretability made it the optimal choice for this classification task as shown in Table~\ref{tab:tab3} and Figure~\ref{fig:algo_select}.
\paragraph*{Neural Network Analysis}
In addition to traditional ML methods, Long Short-Term Memory (LSTM) networks and Convolutional Neural Networks (CNNs) were tested. CNNs achieved the higher F1-micro score of 0.74544, demonstrating their potential for capturing complex hierarchical patterns in the function call data. 
\paragraph*{Robustness Testing}
To ensure the reliability and generalizability of the multi-label classification model, robustness testing was performed. The balanced dataset was subjected to different levels of noise by perturbing the feature values with Gaussian noise ($\mu = 0$, $\sigma = 0.01$ and $\sigma = 0.05$). The XGBoost model demonstrated strong resilience, with a marginal reduction in F1-macro and micro scores by an average of 0.02 and 0.01, respectively, even under the highest noise level. Additionally, the model was evaluated on an imbalanced subset of the original dataset to verify its ability to adapt to real-world class distributions. While the performance dropped slightly, with an F1-macro score of 0.839 and an F1-micro score of 0.852, the model remained effective, indicating its robustness against both noise and distributional shifts.

\paragraph*{Final Model and Results}
After hyperparameter optimization using \texttt{RandomizedSearchCV}, the XGBoost model was retrained, reaching final F1-macro and micro scores of 0.858 and 0.867, respectively, on the test set. This confirmed the model’s ability to generalize across unseen data and highlighted the efficacy of the selected features in distinguishing system tasks.

\begin{table}[htbp]
\centering
\caption{Performance Comparison of Algorithms for Multi-Label Classification}
\label{tab:tab3}
\begin{tabular}{|l|c|c|}
\hline
\textbf{Algorithm}         & \textbf{F1-Macro} & \textbf{F1-Micro} \\ \hline
Random Forest              & 0.812             & 0.824             \\ \hline
Gradient Boosting          & 0.831             & 0.840             \\ \hline
Logistic Regression        & 0.785             & 0.798             \\ \hline
Support Vector Machines    & 0.810             & 0.820             \\ \hline
Multi-Layer Perceptron     & 0.828             & 0.838             \\ \hline
XGBoost                    & 0.858             & 0.867             \\ \hline

\end{tabular}
\end{table}
\section{Conclusion and Future Research Directions}

This study highlights the effectiveness of the Linux kernel’s \texttt{ftrace} framework, particularly the \texttt{function\_graph} tracer, in augmenting ML-based system behavior analysis. By deriving structural and temporal features from function call graphs, we achieved 99.28\% accuracy in binary classification and strong results in multi-label tasks, demonstrating the value of combining system-level tracing with ML techniques.

\subsection{Challenges and Limitations}
Key challenges included managing high-dimensional feature spaces and the computational cost of large-scale graph processing. Moreover, while \texttt{function\_graph} offers rich insights, its runtime overhead limits scalability for real-time use.

\subsection{Future Research Directions}
Future work may expand on several fronts:
\begin{itemize}
    \item \textbf{Ransomware Detection}: Adapting the methodology for real-time identification of ransomware encryption via selective tracing and optimized event pipelines to reduce latency while retaining accuracy. 
    \item \textbf{Application Whitelisting}: Using multi-label classification to validate application behavior against operational baselines and detect unauthorized modifications. 
    \item \textbf{Temporal Graph Analysis}: Employing dynamic GNNs or recurrent models to study evolving function call graphs. 
    \item \textbf{Advanced Feature Representation}: Leveraging GNNs or attention-based models to better capture structural dependencies in call graphs. 
    \item \textbf{Real-time Lightweight Tracing}: Optimizing \texttt{ftrace} with adaptive sampling or preprocessing pipelines for live monitoring. 
    \item \textbf{Extended Security Applications}: Applying this approach to anomaly detection, insider threat mitigation, and proactive defense using additional system-level data. 
    \item \textbf{Cross-Platform Adaptation}: Extending the methodology to other OS-level tracers such as ETW for Windows~\cite{etw_overview} and DTrace for macOS/BSD~\cite{dtrace_overview}.
\end{itemize}

\subsection{Final Remarks}
This work demonstrates how kernel-level tracing integrated with ML can enhance security and system analysis. Addressing the identified challenges and extending research directions can further advance ransomware detection, application whitelisting, and system defense using \texttt{ftrace}-based methodologies.

\end{document}